\pdfoutput=1

\documentclass[11pt]{article}

\usepackage[preprint]{acl}

\usepackage{times}
\usepackage{latexsym}

\usepackage[T1]{fontenc}

\usepackage[utf8]{inputenc}

\usepackage{microtype}

\usepackage{inconsolata}

\usepackage{graphicx}

\usepackage{tabularray}
\usepackage{subfigure}

\title{Memorization in Language Models through the Lens of Intrinsic Dimension}

\author{Stefan Arnold \\ 
Friedrich-Alexander-Universität Erlangen-Nürnberg \\
Lange Gasse 20, 90403 Nürnberg, Germany \\ 
\texttt{stefan.st.arnold@fau.de}}

\begin{document}
\maketitle
\begin{abstract}

\textit{Language Models} (LMs) are prone to memorizing parts of their data during training and unintentionally emitting them at generation time, raising concerns about privacy leakage and disclosure of intellectual property. While previous research has identified properties such as context length, parameter size, and duplication frequency, as key drivers of unintended memorization, little is known about how the latent structure modulates this rate of memorization. We investigate the role of \textit{Intrinsic Dimension} (ID), a geometric proxy for the structural complexity of a sequence in latent space, in modulating memorization. Our findings suggest that ID acts as a suppressive signal for memorization: compared to low-ID sequences, high-ID sequences are less likely to be memorized, particularly in overparameterized models and under sparse exposure. These findings highlight the interaction between scale, exposure, and complexity in shaping memorization.

\end{abstract}

\section{Introduction}

\textit{Language Models} (LMs) \citep{brown2020language, raffel2020exploring, chowdhery2023palm} are susceptible to memorizing segments of texts encountered during training \citep{shokri2017membership} and emitting these segments during generation \citep{nasr2025scalable}, even from corpora that has been subjected to deduplication \citep{kandpal2022deduplicating, lee2022deduplicating}. While memorization is connected to generalization \citep{arpit2017closer, brown2021memorization}, it can cause severe issues such as inadvertent reproduction of personal information \citep{huang2022large} and copyrighted materials \citep{lee2023language}. 

\begin{figure}[!tb]
    \centering
    \includegraphics[width=0.45\textwidth]{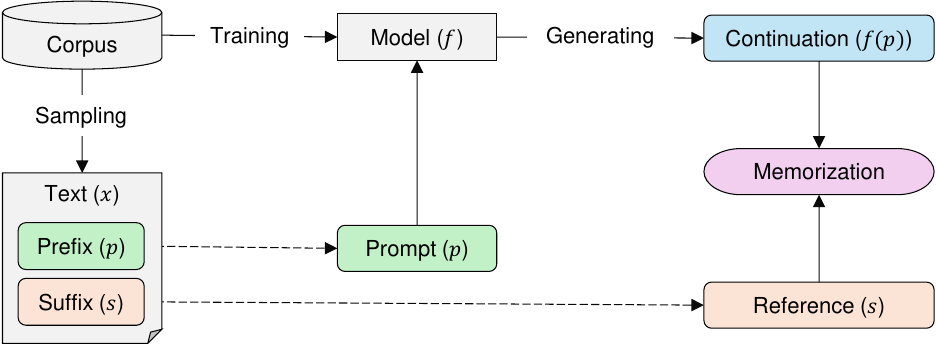}
    \caption{Overview of the post-hoc assessment of memorization, adapted from \citet{kiyomaru2024comprehensive}. Methodologically, a sample $x$ is split into a prefix $p$ and a suffix $s$. By prompting $p$, the model $f$ generates a continuation $f(p)$. If the continuation $f(p)$ matches $s$ verbatim, the instance $x$ is considered memorized.}
    \label{fig:approach}
\end{figure}

To estimate memorization rates of LMs, \citet{carlini2019secret} formalized a loose bound on memorization known as \textit{exposure}, a metric that measures the relative difference in log-perplexity between \textit{canaries}, synthetic sequences of text with fixed formats that are inserted during training and extracted during generation. By leveraging examples directly from the corpus, \citet{carlini2023quantifying} introduced a tighter bound on memorization that avoids the need for canaries and reduces the computational overhead associated with computing exposure. Figure \ref{fig:approach} visualizes the actionable methodology for examining memorization. Given a subset of examples, each split into a prefix $p$ and a suffix $s$, memorization is estimated post-hoc by prompting the model $f$ with the prefix and checking whether its continuation $f(p)$ replicates the reference $s$. The proportion of continuations that match the references verbatim provides an empirical estimate of memorization and quantifies the risk of information leakage. 

Once memorization was evidenced in practice \citep{nasr2023scalable}, several properties have been identified as factors contributing to the memorization rate. Beyond its correlation with overfitting \citep{yeom2018privacy}, memorization is related to duplication counts \citep{carlini2023quantifying, ippolito2023preventing, zhang2023counterfactual, kiyomaru2024comprehensive}, model capacity \citep{tirumala2022memorization, carlini2023quantifying}, and context length \citep{carlini2023quantifying}. 

Grounded on the manifold hypothesis \cite{fefferman2016testing}, few studies have examined the intrinsic dimension of data representations as a means to understand how neural networks structure latent spaces. These studies reveal that high-dimensional signals tend to lie in low-dimensional subspaces \citep{ansuini2019intrinsic}, and that intrinsic dimensionality acts as a geometric proxy for generalization capacity \citep{birdal2021intrinsic, pope2021intrinsic}.

\paragraph{Contribution.} Assuming that the intrinsic dimension offers a lens onto sample complexity of sequences \textit{as perceived} by language models, we investigate its relationship to the likelihood of memorization. Our investigation reveals that the intrinsic dimension systematically modulates memorization behavior: sequences with low intrinsic dimension, residing in compressed subspaces, are more amenable to memorization, particularly under sparse exposure, whereas sequences with high intrinsic dimension are less frequently memorized unless they are encountered repeatedly.


\section{Background}
\label{sec:background}

We briefly provide necessary foundations for unintended memorization and intrinsic dimensionality.

\subsection{Unintended Memorization}

Memorization is commonly referred to the phenomenon of a neural network to fit arbitrarily assigned labels to features \citep{zhang2022understanding}. Although viewed as a sign of overfitting, memorization is linked to generalization \citep{arpit2017closer}, particularly for data with long-tailed distributions \citep{feldman2020does, feldman2020neural}, where memorization can serve as an inductive bias that enables models to generalize beyond dominant modes and learn from rare or noisy examples.

\textit{Unintended Memorization}, which refers to the reproduction of data used for training during generation, stands in contrast to these desirable forms of memorization \citep{brown2021memorization}. A longstanding belief held that memorization arises in the presence of overfitting \citep{yeom2018privacy}, however, this belief has been challenged by recent findings showing memorization in the absence of overfitting \citep{tirumala2022memorization}. Since large-scale language models have been found to memorize content even when trained on massively deduplicated text, overfitting only presents a sufficient condition but not a necessary condition for memorization. 

Calling for a more nuanced understanding of unintended memorization, several notions have been operationalized. Depending on their degree of fidelity, these notions can be broadly categorized into \textit{verbatim memorization}, in which sequences must match exactly, and \textit{approximate memorization}, which allows for slight variations \citep{ippolito2023preventing}. Noteable definitions for memorization include \textit{canary memorization} \citep{carlini2019secret}, \textit{eidetic memorization} \citep{carlini2021extracting}, \textit{counterfactual memorization} \citep{feldman2020neural, zhang2023counterfactual}, \textit{discoverable memorization} \citep{carlini2023quantifying, hayes2024measuring}, and \textit{distributional memorization} \citep{wang2025generalization}. 

We adopt discoverable memorization as our actionable notion of memorization, formalizing the scenario in which a language model is prompted with the prefix of an example and is deemed to have memorized it if its continuation reproduces the suffix of the example \textit{verbatim}. \citet{carlini2023quantifying} operationalize this definition using deterministic decoding via greedy sampling, whereas \citet{hayes2024measuring} demonstrate its robustness across decoding strategies by accounting for temperature sampling.

\subsection{Intrinsic Dimensionality}

Unlike the ambient dimension of a representation space, the notion of \textit{Intrinsic Dimension}  (ID) characterizes the minimum number of latent directions required to represent data with minimal information loss \citep{fefferman2016testing}. Geometrically, ID describes the manifold on which the data points are concentrated, capturing the effective dimensionality. The ID property has been used to gain insight into the sequential information flow in neural networks. \citet{ansuini2019intrinsic} showed that neural networks progressively compress high-dimensional data into low-dimensional manifolds, forming representations with orders-of-magnitude lower dimensionality than the ambient space.


A prototypical approach to estimate the ID involves projecting data onto a linear subspace \citep{jolliffe1986mathematical}. Since techniques relying on a linear projection poorly estimate the ID for data lying on curved manifolds, more recent techniques exploit local structures from nearest neighbors \citep{levina2004maximum, farahmand2007manifold, facco2017estimating, amsaleg2018extreme} or leverage the global topology \citep{schweinhart2021persistent}.

\citet{levina2004maximum} uses maximum likelihood estimation to fit a likelihood on the distances from a given point to its $k$-nearest points within a neighborhood structure. To stabilize ID estimations when confronted with variations in densities and curvatures within a manifold, \citet{facco2017estimating} considers only the ratio of distances between two closest neighbors, providing robust estimation from minimal neighborhood information. \citet{schweinhart2021persistent} recently connects ID estimation to the well-established field of persistent homology by characterizing the continuous shape of the manifold at different scales to the upper box dimension. The upper box dimension is related to how efficiently points can be covered by boxes of decreasing size.

\section{Methodology}

We build on the setup introduced by \citet{carlini2023quantifying} to assess memorization in relation to structural complexity. Specifically, we employ the \texttt{GPT}-neo model family \citep{wang2021gpt} and reuse their random sample derived from the \texttt{Pile} \citep{gao2020pile}. To ensure that our measurements isolate structural complexity from confounding factors, we carefully control sequence length and duplication counts. We restrict all sequences to a uniform length of $150$, thereby stabilizing ID estimations. We subsample $1,000$ sequences stratified by duplication frequency on a logarithmic scale for ranges between $[1,10)$, $[10,100)$, and $[100,1000)$, allowing us to disentangle the influence of duplication from that of structural complexity.

\begin{table}
\centering
\caption{Examples of text and their corresponding number of dimensions occupied in latent space. Higher ID values indicate greater geometric complexity.}
\label{tab:examples}
\begin{tblr}{
  width = \linewidth,
  colspec = {Q[906]Q[37]},
  column{2} = {r},
  hline{1,4} = {-}{0.08em},
  hline{2} = {-}{0.05em},
}
\small \textbf{Text} (truncated) & \small \textbf{ID} \\
\small We shall have no responsibility or liability for your visitation to, and the data collection and use practices of, such other sites. This Policy applies solely to the information collected in connection with your use of this Website and does not apply to any practices conducted offline or in connection with any other websites. [...] & \small 2.08 \\
\small Kazuni area there are many hippos and crocodiles which although rarely seen from the shore can certainly be heard at night. The location of the small town of Nkhata Bay is quite spectacular, a large, sheltered bay, accessible via a steep slope. Small boats transport the local people to various locations so that they can buy and sell, as there are hardly any roads around the lake. [...] & \small 9.07        
\end{tblr}
\end{table}

To estimate the ID, we follow \citet{tulchinskii2024intrinsic} by treating each text as a point cloud spanning a manifold in the embedding space. We then obtain contextualized embeddings using \texttt{BERT} \citep{devlin2019bert}, and estimate the intrinsic dimension using \texttt{TwoNN} \citep{facco2017estimating}, discarding artifacts of tokenization. Table~\ref{tab:examples} depicts example sequences and their corresponding IDs, which we interpret as a proxy for complexity in latent space.

\begin{figure}[!tb]
    \centering
    \includegraphics[width=0.40\textwidth]{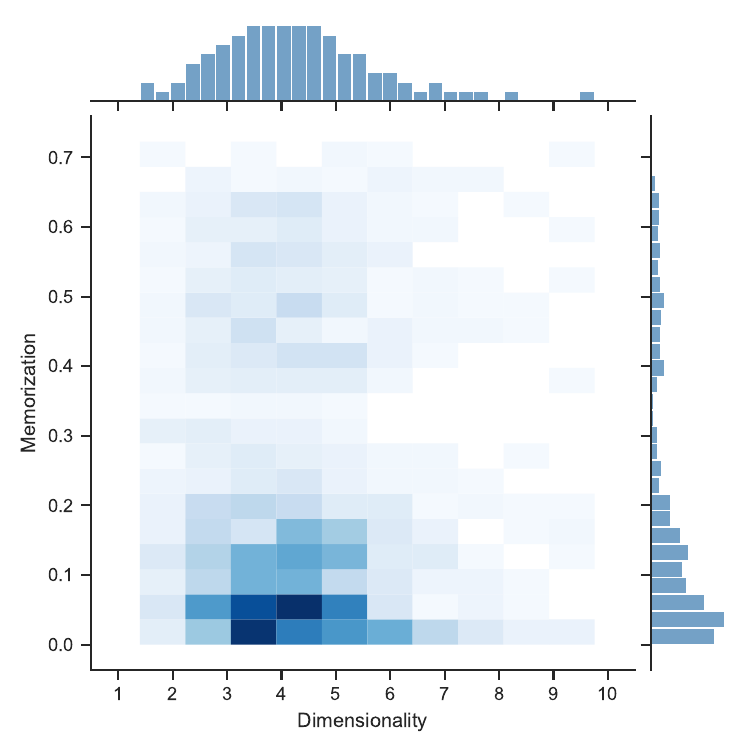}
    \caption{Distribution of memorization rate and intrinsic dimension, aggregated across scale and exposure.}
    \label{fig:histogram}
\end{figure}

Figure~\ref{fig:histogram} shows the joint distribution of the memorization rate and intrinsic dimensionality, aggregated across model sizes and duplication counts. We observe that most samples cluster in regions characterized by low dimensionality and low memorization. However, when disaggregating by model scale and number of duplications, clear patterns emerge that elucidate the relationship between structural complexity and rate of memorization.


\section{Findings}

\begin{figure*}[t]
    \centering
    \subfigure[$[1,10)$]
    {
        \centering
        \includegraphics[width=0.3\textwidth]{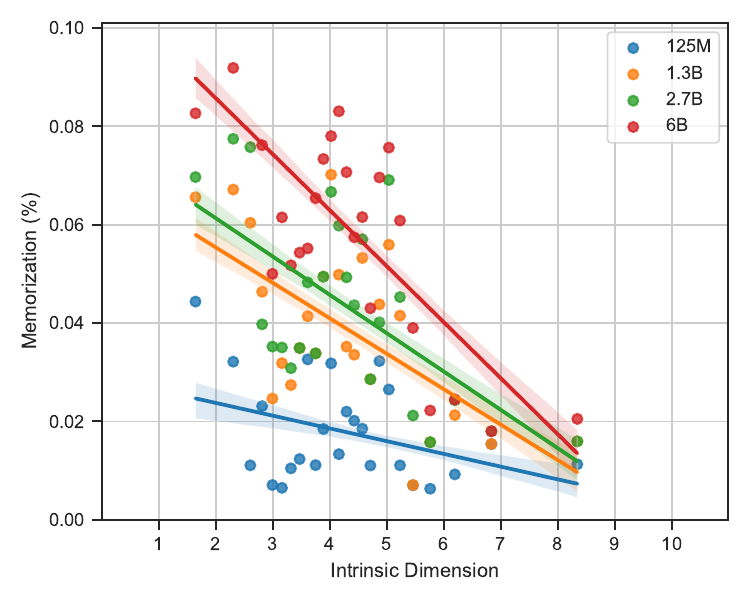}
        \label{fig:1_10}
    }
    \subfigure[$[10,100)$]
    {
        \centering
        \includegraphics[width=0.3\textwidth]{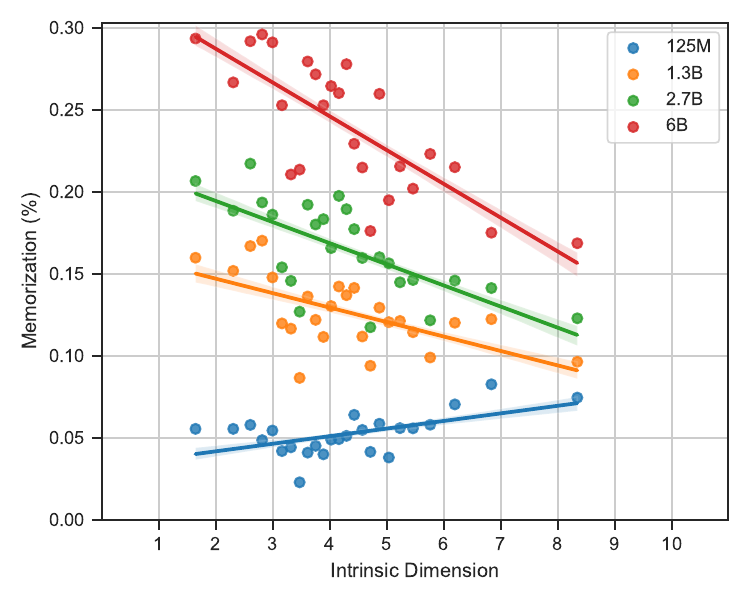}
        \label{fig:10_100}
    }
    \subfigure[$[100,1000)$]
    {
        \centering
        \includegraphics[width=0.3\textwidth]{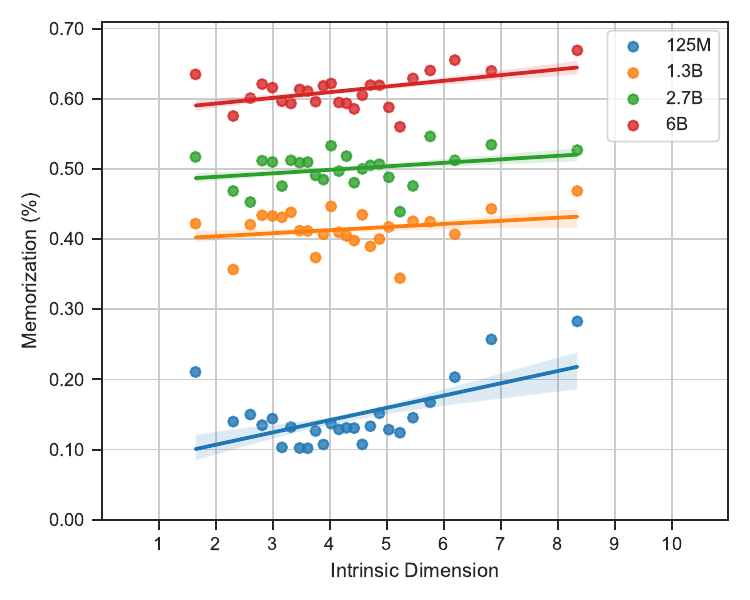}
        \label{fig:100_1000}
    }
    \caption{Memorization as a function of intrinsic memorization, binned into equally-sized intervals and disaggregated by model scale. \ref{fig:1_10} presents a low-duplication regime, comprising samples with duplications of at most $10$. \ref{fig:10_100} presents a medium-duplication regime, comprising samples with duplication frequencies ranging from $10$ to $100$. \ref{fig:100_1000} presents a high-duplication regime, comprising samples with duplications capped at $1000$.}
    \label{fig:manifold}
\end{figure*}

Figure~\ref{fig:manifold} presents the relationship between memorization rate and intrinsic dimensionality for ascending levels of duplication frequency. Specifically, we quantile-binned the intrinsic dimension into $25$ equally sized intervals and averaged memorization within each bin.  Each subplot further disaggregates model capacity, covering models with roughly $0.1$, $1.3$, $2.7$, and $6.0$ billion  parameters.

Consistent with the relationships reported by \citet{carlini2023quantifying}, our findings reveal a log-linear increase of memorization as a function of both duplication count and model capacity. Beyond these relationships, we observe a modulating influence of the intrinsic dimension. In the low-duplication regime, memorization declines inversely with intrinsic dimensionality across all model sizes. This inverse trend indicates that complex sequences, particularly those lying on more intricate manifolds, are less likely to be memorized under sparse exposure. In the medium-duplication regime, we notice diverging patterns depending on the model sizes. The inverse relationship largely persists for large models, albeit with a diminished effect. However, this is not the case for small models. Once duplications are sufficiently frequent for memorization, small models display a reversal in trend, exhibiting a slight increase in memorization with rising structural complexity. This divergence may reflect the limited capacity of certain models to generalize, leading to greater memorization of sequences that they fail to compress effectively. In the high-duplication regime, memorization undergoes a further shift as it saturates and becomes almost invariant to the intrinsic dimension. These findings suggest that under conditions of frequent exposure, memorization is increasingly governed by exposure and scale, overriding the modulating influence of structural complexity.

\section{Conclusion}

Building on the shared connection of memorization and intrinsic dimension to generalization, we introduce the intrinsic dimension as a complementary factor shaping the likelihood of memorization in language models. Specifically, we examine the relationship between memorization rate and the structural complexity of sequences in latent space, conditioned on model scale and exposure frequency. For sufficiently parameterized models and moderate levels of duplication, the intrinsic dimension act as a suppressive signal on memorization. A reversed trend can be seen for models with limited capacity, which tend to memorize structurally complex sequences even under moderate exposure.

\paragraph{Limitations.}  Despite controlling for duplication frequency, we focus exclusively on exact duplicates, omitting near-duplicates which are known to account for the majority of memorized content in large-scale corpora \citep{lee2022deduplicating}. This constraint likely underestimates  memorization. Additionally, we restrict our analysis to verbatim memorization, a narrow definition that is known to give a false sense of privacy \citep{ippolito2023preventing}. Finally, we rely on greedy decoding to measure memorization, however, this decoding strategy is atypical in practical deployments \citep{hayes2024measuring}.

\bibliography{custom}

\begin{thebibliography}{39}
\providecommand{\natexlab}[1]{#1}

\bibitem[{Amsaleg et~al.(2018)Amsaleg, Chelly, Furon, Girard, Houle, Kawarabayashi, and Nett}]{amsaleg2018extreme}
Laurent Amsaleg, Oussama Chelly, Teddy Furon, St{\'e}phane Girard, Michael~E Houle, Ken-ichi Kawarabayashi, and Michael Nett. 2018.
\newblock Extreme-value-theoretic estimation of local intrinsic dimensionality.
\newblock \emph{Data Mining and Knowledge Discovery}, 32(6):1768--1805.

\bibitem[{Ansuini et~al.(2019)Ansuini, Laio, Macke, and Zoccolan}]{ansuini2019intrinsic}
Alessio Ansuini, Alessandro Laio, Jakob~H Macke, and Davide Zoccolan. 2019.
\newblock Intrinsic dimension of data representations in deep neural networks.
\newblock \emph{Advances in Neural Information Processing Systems}, 32.

\bibitem[{Arpit et~al.(2017)Arpit, Jastrz{\k{e}}bski, Ballas, Krueger, Bengio, Kanwal, Maharaj, Fischer, Courville, Bengio et~al.}]{arpit2017closer}
Devansh Arpit, Stanis{\l}aw Jastrz{\k{e}}bski, Nicolas Ballas, David Krueger, Emmanuel Bengio, Maxinder~S Kanwal, Tegan Maharaj, Asja Fischer, Aaron Courville, Yoshua Bengio, et~al. 2017.
\newblock A closer look at memorization in deep networks.
\newblock In \emph{International conference on machine learning}, pages 233--242. PMLR.

\bibitem[{Birdal et~al.(2021)Birdal, Lou, Guibas, and Simsekli}]{birdal2021intrinsic}
Tolga Birdal, Aaron Lou, Leonidas~J Guibas, and Umut Simsekli. 2021.
\newblock Intrinsic dimension, persistent homology and generalization in neural networks.
\newblock \emph{Advances in Neural Information Processing Systems}, 34:6776--6789.

\bibitem[{Brown et~al.(2021)Brown, Bun, Feldman, Smith, and Talwar}]{brown2021memorization}
Gavin Brown, Mark Bun, Vitaly Feldman, Adam Smith, and Kunal Talwar. 2021.
\newblock When is memorization of irrelevant training data necessary for high-accuracy learning?
\newblock In \emph{Proceedings of the 53rd annual ACM SIGACT symposium on theory of computing}, pages 123--132.

\bibitem[{Brown et~al.(2020)Brown, Mann, Ryder, Subbiah, Kaplan, Dhariwal, Neelakantan, Shyam, Sastry, Askell et~al.}]{brown2020language}
Tom Brown, Benjamin Mann, Nick Ryder, Melanie Subbiah, Jared~D Kaplan, Prafulla Dhariwal, Arvind Neelakantan, Pranav Shyam, Girish Sastry, Amanda Askell, et~al. 2020.
\newblock Language models are few-shot learners.
\newblock \emph{Advances in neural information processing systems}, 33:1877--1901.

\bibitem[{Carlini et~al.(2023)Carlini, Ippolito, Jagielski, Lee, Tramer, and Zhang}]{carlini2023quantifying}
Nicholas Carlini, Daphne Ippolito, Matthew Jagielski, Katherine Lee, Florian Tramer, and Chiyuan Zhang. 2023.
\newblock Quantifying memorization across neural language models.
\newblock In \emph{The Eleventh International Conference on Learning Representations}.

\bibitem[{Carlini et~al.(2019)Carlini, Liu, Erlingsson, Kos, and Song}]{carlini2019secret}
Nicholas Carlini, Chang Liu, {\'U}lfar Erlingsson, Jernej Kos, and Dawn Song. 2019.
\newblock The secret sharer: Evaluating and testing unintended memorization in neural networks.
\newblock In \emph{28th USENIX Security Symposium (USENIX Security 19)}, pages 267--284.

\bibitem[{Carlini et~al.(2021)Carlini, Tramer, Wallace, Jagielski, Herbert-Voss, Lee, Roberts, Brown, Song, Erlingsson et~al.}]{carlini2021extracting}
Nicholas Carlini, Florian Tramer, Eric Wallace, Matthew Jagielski, Ariel Herbert-Voss, Katherine Lee, Adam Roberts, Tom Brown, Dawn Song, Ulfar Erlingsson, et~al. 2021.
\newblock Extracting training data from large language models.
\newblock In \emph{30th USENIX security symposium (USENIX Security 21)}, pages 2633--2650.

\bibitem[{Chowdhery et~al.(2023)Chowdhery, Narang, Devlin, Bosma, Mishra, Roberts, Barham, Chung, Sutton, Gehrmann et~al.}]{chowdhery2023palm}
Aakanksha Chowdhery, Sharan Narang, Jacob Devlin, Maarten Bosma, Gaurav Mishra, Adam Roberts, Paul Barham, Hyung~Won Chung, Charles Sutton, Sebastian Gehrmann, et~al. 2023.
\newblock Palm: Scaling language modeling with pathways.
\newblock \emph{Journal of Machine Learning Research}, 24(240):1--113.

\bibitem[{Devlin et~al.(2019)Devlin, Chang, Lee, and Toutanova}]{devlin2019bert}
Jacob Devlin, Ming-Wei Chang, Kenton Lee, and Kristina Toutanova. 2019.
\newblock \href {https://doi.org/10.18653/v1/N19-1423} {{BERT}: Pre-training of deep bidirectional transformers for language understanding}.
\newblock In \emph{Proceedings of the 2019 Conference of the North {A}merican Chapter of the Association for Computational Linguistics: Human Language Technologies, Volume 1 (Long and Short Papers)}, pages 4171--4186, Minneapolis, Minnesota. Association for Computational Linguistics.

\bibitem[{Facco et~al.(2017)Facco, d’Errico, Rodriguez, and Laio}]{facco2017estimating}
Elena Facco, Maria d’Errico, Alex Rodriguez, and Alessandro Laio. 2017.
\newblock Estimating the intrinsic dimension of datasets by a minimal neighborhood information.
\newblock \emph{Scientific reports}, 7(1):12140.

\bibitem[{Farahmand et~al.(2007)Farahmand, Szepesv{\'a}ri, and Audibert}]{farahmand2007manifold}
Amir~Massoud Farahmand, Csaba Szepesv{\'a}ri, and Jean-Yves Audibert. 2007.
\newblock Manifold-adaptive dimension estimation.
\newblock In \emph{Proceedings of the 24th international conference on Machine learning}, pages 265--272.

\bibitem[{Fefferman et~al.(2016)Fefferman, Mitter, and Narayanan}]{fefferman2016testing}
Charles Fefferman, Sanjoy Mitter, and Hariharan Narayanan. 2016.
\newblock Testing the manifold hypothesis.
\newblock \emph{Journal of the American Mathematical Society}, 29(4):983--1049.

\bibitem[{Feldman(2020)}]{feldman2020does}
Vitaly Feldman. 2020.
\newblock Does learning require memorization? a short tale about a long tail.
\newblock In \emph{Proceedings of the 52nd Annual ACM SIGACT Symposium on Theory of Computing}, pages 954--959.

\bibitem[{Feldman and Zhang(2020)}]{feldman2020neural}
Vitaly Feldman and Chiyuan Zhang. 2020.
\newblock What neural networks memorize and why: Discovering the long tail via influence estimation.
\newblock \emph{Advances in Neural Information Processing Systems}, 33:2881--2891.

\bibitem[{Gao et~al.(2020)Gao, Biderman, Black, Golding, Hoppe, Foster, Phang, He, Thite, Nabeshima et~al.}]{gao2020pile}
Leo Gao, Stella Biderman, Sid Black, Laurence Golding, Travis Hoppe, Charles Foster, Jason Phang, Horace He, Anish Thite, Noa Nabeshima, et~al. 2020.
\newblock The pile: An 800gb dataset of diverse text for language modeling.
\newblock \emph{arXiv preprint arXiv:2101.00027}.

\bibitem[{Hayes et~al.(2024)Hayes, Swanberg, Chaudhari, Yona, and Shumailov}]{hayes2024measuring}
Jamie Hayes, Marika Swanberg, Harsh Chaudhari, Itay Yona, and Ilia Shumailov. 2024.
\newblock Measuring memorization through probabilistic discoverable extraction.
\newblock \emph{arXiv preprint arXiv:2410.19482}.

\bibitem[{Huang et~al.(2022)Huang, Shao, and Chang}]{huang2022large}
Jie Huang, Hanyin Shao, and Kevin Chen-Chuan Chang. 2022.
\newblock \href {https://doi.org/10.18653/v1/2022.findings-emnlp.148} {Are large pre-trained language models leaking your personal information?}
\newblock In \emph{Findings of the Association for Computational Linguistics: EMNLP 2022}, pages 2038--2047, Abu Dhabi, United Arab Emirates. Association for Computational Linguistics.

\bibitem[{Ippolito et~al.(2023)Ippolito, Tramer, Nasr, Zhang, Jagielski, Lee, Choo, and Carlini}]{ippolito2023preventing}
Daphne Ippolito, Florian Tramer, Milad Nasr, Chiyuan Zhang, Matthew Jagielski, Katherine Lee, Christopher~Choquette Choo, and Nicholas Carlini. 2023.
\newblock Preventing generation of verbatim memorization in language models gives a false sense of privacy.
\newblock In \emph{Proceedings of the 16th International Natural Language Generation Conference}, pages 28--53.

\bibitem[{Jolliffe and Jolliffe(1986)}]{jolliffe1986mathematical}
Ian~T Jolliffe and IT~Jolliffe. 1986.
\newblock \emph{Mathematical and statistical properties of sample principal components}.
\newblock Springer.

\bibitem[{Kandpal et~al.(2022)Kandpal, Wallace, and Raffel}]{kandpal2022deduplicating}
Nikhil Kandpal, Eric Wallace, and Colin Raffel. 2022.
\newblock Deduplicating training data mitigates privacy risks in language models.
\newblock In \emph{International Conference on Machine Learning}, pages 10697--10707. PMLR.

\bibitem[{Kiyomaru et~al.(2024)Kiyomaru, Sugiura, Kawahara, and Kurohashi}]{kiyomaru2024comprehensive}
Hirokazu Kiyomaru, Issa Sugiura, Daisuke Kawahara, and Sadao Kurohashi. 2024.
\newblock A comprehensive analysis of memorization in large language models.
\newblock In \emph{Proceedings of the 17th International Natural Language Generation Conference}, pages 584--596.

\bibitem[{Lee et~al.(2023)Lee, Le, Chen, and Lee}]{lee2023language}
Jooyoung Lee, Thai Le, Jinghui Chen, and Dongwon Lee. 2023.
\newblock Do language models plagiarize?
\newblock In \emph{Proceedings of the ACM Web Conference 2023}, pages 3637--3647.

\bibitem[{Lee et~al.(2022)Lee, Ippolito, Nystrom, Zhang, Eck, Callison-Burch, and Carlini}]{lee2022deduplicating}
Katherine Lee, Daphne Ippolito, Andrew Nystrom, Chiyuan Zhang, Douglas Eck, Chris Callison-Burch, and Nicholas Carlini. 2022.
\newblock \href {https://doi.org/10.18653/v1/2022.acl-long.577} {Deduplicating training data makes language models better}.
\newblock In \emph{Proceedings of the 60th Annual Meeting of the Association for Computational Linguistics (Volume 1: Long Papers)}, pages 8424--8445, Dublin, Ireland. Association for Computational Linguistics.

\bibitem[{Levina and Bickel(2004)}]{levina2004maximum}
Elizaveta Levina and Peter Bickel. 2004.
\newblock Maximum likelihood estimation of intrinsic dimension.
\newblock \emph{Advances in neural information processing systems}, 17.

\bibitem[{Nasr et~al.(2023)Nasr, Carlini, Hayase, Jagielski, Cooper, Ippolito, Choquette-Choo, Wallace, Tram{\`e}r, and Lee}]{nasr2023scalable}
Milad Nasr, Nicholas Carlini, Jonathan Hayase, Matthew Jagielski, A~Feder Cooper, Daphne Ippolito, Christopher~A Choquette-Choo, Eric Wallace, Florian Tram{\`e}r, and Katherine Lee. 2023.
\newblock Scalable extraction of training data from (production) language models.
\newblock \emph{arXiv preprint arXiv:2311.17035}.

\bibitem[{Nasr et~al.(2025)Nasr, Rando, Carlini, Hayase, Jagielski, Cooper, Ippolito, Choquette-Choo, Tram{\`e}r, and Lee}]{nasr2025scalable}
Milad Nasr, Javier Rando, Nicholas Carlini, Jonathan Hayase, Matthew Jagielski, A~Feder Cooper, Daphne Ippolito, Christopher~A Choquette-Choo, Florian Tram{\`e}r, and Katherine Lee. 2025.
\newblock Scalable extraction of training data from aligned, production language models.
\newblock In \emph{The Thirteenth International Conference on Learning Representations}.

\bibitem[{Pope et~al.(2021)Pope, Zhu, Abdelkader, Goldblum, and Goldstein}]{pope2021intrinsic}
Phillip Pope, Chen Zhu, Ahmed Abdelkader, Micah Goldblum, and Tom Goldstein. 2021.
\newblock The intrinsic dimension of images and its impact on learning.
\newblock \emph{9th International Conference on Learning Representations, {ICLR}}.

\bibitem[{Raffel et~al.(2020)Raffel, Shazeer, Roberts, Lee, Narang, Matena, Zhou, Li, and Liu}]{raffel2020exploring}
Colin Raffel, Noam Shazeer, Adam Roberts, Katherine Lee, Sharan Narang, Michael Matena, Yanqi Zhou, Wei Li, and Peter~J Liu. 2020.
\newblock Exploring the limits of transfer learning with a unified text-to-text transformer.
\newblock \emph{Journal of machine learning research}, 21(140):1--67.

\bibitem[{Schweinhart(2021)}]{schweinhart2021persistent}
Benjamin Schweinhart. 2021.
\newblock Persistent homology and the upper box dimension.
\newblock \emph{Discrete \& Computational Geometry}, 65(2):331--364.

\bibitem[{Shokri et~al.(2017)Shokri, Stronati, Song, and Shmatikov}]{shokri2017membership}
Reza Shokri, Marco Stronati, Congzheng Song, and Vitaly Shmatikov. 2017.
\newblock Membership inference attacks against machine learning models.
\newblock In \emph{2017 IEEE symposium on security and privacy (SP)}, pages 3--18. IEEE.

\bibitem[{Tirumala et~al.(2022)Tirumala, Markosyan, Zettlemoyer, and Aghajanyan}]{tirumala2022memorization}
Kushal Tirumala, Aram Markosyan, Luke Zettlemoyer, and Armen Aghajanyan. 2022.
\newblock Memorization without overfitting: Analyzing the training dynamics of large language models.
\newblock \emph{Advances in Neural Information Processing Systems}, 35:38274--38290.

\bibitem[{Tulchinskii et~al.(2024)Tulchinskii, Kuznetsov, Kushnareva, Cherniavskii, Nikolenko, Burnaev, Barannikov, and Piontkovskaya}]{tulchinskii2024intrinsic}
Eduard Tulchinskii, Kristian Kuznetsov, Laida Kushnareva, Daniil Cherniavskii, Sergey Nikolenko, Evgeny Burnaev, Serguei Barannikov, and Irina Piontkovskaya. 2024.
\newblock Intrinsic dimension estimation for robust detection of ai-generated texts.
\newblock \emph{Advances in Neural Information Processing Systems}, 36.

\bibitem[{Wang and Komatsuzaki(2021)}]{wang2021gpt}
Ben Wang and Aran Komatsuzaki. 2021.
\newblock Gpt-j-6b: A 6 billion parameter autoregressive language model.

\bibitem[{Wang et~al.(2025)Wang, Antoniades, Elazar, Amayuelas, Albalak, Zhang, and Wang}]{wang2025generalization}
Xinyi Wang, Antonis Antoniades, Yanai Elazar, Alfonso Amayuelas, Alon Albalak, Kexun Zhang, and William~Yang Wang. 2025.
\newblock Generalization v.s. memorization: Tracing language models{\textquoteright} capabilities back to pretraining data.
\newblock In \emph{The Thirteenth International Conference on Learning Representations}.

\bibitem[{Yeom et~al.(2018)Yeom, Giacomelli, Fredrikson, and Jha}]{yeom2018privacy}
Samuel Yeom, Irene Giacomelli, Matt Fredrikson, and Somesh Jha. 2018.
\newblock Privacy risk in machine learning: Analyzing the connection to overfitting.
\newblock In \emph{2018 IEEE 31st computer security foundations symposium (CSF)}, pages 268--282. IEEE.

\bibitem[{Zhang et~al.(2022)Zhang, Bengio, Hardt, Recht, and Vinyals}]{zhang2022understanding}
Chiyuan Zhang, Samy Bengio, Moritz Hardt, Benjamin Recht, and Oriol Vinyals. 2022.
\newblock Understanding deep learning requires rethinking generalization.
\newblock In \emph{International Conference on Learning Representations}.

\bibitem[{Zhang et~al.(2023)Zhang, Ippolito, Lee, Jagielski, Tram{\`e}r, and Carlini}]{zhang2023counterfactual}
Chiyuan Zhang, Daphne Ippolito, Katherine Lee, Matthew Jagielski, Florian Tram{\`e}r, and Nicholas Carlini. 2023.
\newblock Counterfactual memorization in neural language models.
\newblock \emph{Advances in Neural Information Processing Systems}, 36:39321--39362.

\end{thebibliography}
\end{document}